\documentclass{article}

 \usepackage[preprint]{neurips_2026}

\usepackage[utf8]{inputenc} % allow utf-8 input
\usepackage[T1]{fontenc}    % use 8-bit T1 fonts
\usepackage{hyperref}       % hyperlinks
\usepackage{url}            % simple URL typesetting
\usepackage{booktabs}       % professional-quality tables
\usepackage{amsfonts}       % blackboard math symbols
\usepackage{nicefrac}       % compact symbols for 1/2, etc.
\usepackage{microtype}      % microtypography
\usepackage{xcolor}         % colors
\usepackage{amsmath} 
\usepackage{booktabs}
\usepackage{multirow} 
\usepackage{multicol}
\usepackage{subcaption}
\usepackage{longtable}
\usepackage{graphicx}
\usepackage{algorithm}
\usepackage{algorithmic}
\usepackage{amsthm}
\usepackage{amssymb}
\usepackage{pifont}
\usepackage{wrapfig}
\title{Beyond What to Select: A Plug-and-play Oscillatory Data-Volume Scheduling for Efficient Model Training}

\newtheorem{theorem}{Theorem}[section]
\newtheorem{proposition}[theorem]{Proposition}

\newtheorem{corollary}[theorem]{Corollary}

\usepackage{xspace}
\usepackage{soul}
\author{%
Suorong Yang$^{1,2}$ \quad Hanqi Zhu$^{2}$ \quad Hai Gan$^2$ \quad Fangjian Su$^2$ \\
 \textbf{Guang Li}$^3$ \quad \textbf{Furao Shen}*$^2$ \quad \textbf{Soujanya Poria}*$^4$ \\
$^1$National University of Singapore \quad $^2$Nanjing University \quad $^3$Hokkaido University \quad \\ $^4$Nanyang Technological University \\
\texttt{sryang@smail.nju.edu.cn} \quad
\texttt{soujanya.poria@ntu.edu.sg}
}

\newcommand{\model}{\texttt{PODS}}
\begin{document}
% \footnote{* Corresponding authors.}

\maketitle
\begingroup
\renewcommand{\thefootnote}{}
\footnotetext{\textsuperscript{*} Corresponding authors.}
\endgroup

\begin{abstract}
Data selection accelerates training by identifying representative training data while preserving model performance.
However, existing methods mainly focus on designing sample-importance criteria, i.e., deciding what to select, while typically fixing the selected data volume as the target ratio throughout training. 
Thus, they are often dynamic in sample identity but static in data volume.
In this work, we revisit data selection from an optimization perspective and show that selected-data training induces an implicit regularization effect modulated by the instantaneous selection ratio.
This reveals a key trade-off: lower ratios amplify selection-induced regularization, whereas higher ratios preserve data coverage and optimization fidelity.
Motivated by this insight, we propose \model{}, a \underline{\textbf{P}}lug-and-play \underline{\textbf{O}}scillatory \underline{\textbf{D}}ata-volume \underline{\textbf{S}}cheduling framework.
Rather than introducing another sample-scoring metric, \model{} serves as a lightweight module that dynamically schedules how much data to select over training.
Under the target selection ratio, \model{} alternates between low-ratio regularization phases and high-ratio recovery phases to exploit selection-induced regularization without sacrificing optimization stability.
With its lightweight, ratio-level, and task-agnostic design, \model{} is compatible with existing static and dynamic selection methods and broadly applicable across training paradigms.
Experiments across various datasets, architectures, and tasks show that \model{} consistently improves the efficiency-generalization trade-off, e.g., reducing ImageNet-1k training cost by 50\% with improved accuracy and accelerating LLM instruction tuning by over $2\times$ without performance degradation.

\end{abstract}

\section{Introduction}
The rapid growth of modern deep models has made training increasingly expensive, especially when optimization relies on large-scale datasets collected from the open world.
Such datasets are often heterogeneous: many samples are redundant or already well learned, and their utility changes across training stages~\cite{infobatch,clip-selection}.
Consequently, training on massive datasets incurs substantial resource consumption without necessarily yielding proportional performance gains.
Data selection thus offers a promising solution by selecting representative training samples while preserving model performance.

Existing data selection methods can be broadly categorized into static~\cite{dataset_pruning,clip-selection,rho-loss,DUAL,moderate} and dynamic approaches~\cite{infobatch,dynamic_pruning,dynamic_pruning-2,yang2025dynamic}.
Given a target selection ratio, static methods select a fixed subset before training begins, whereas dynamic methods update the selected subsets online as the training state evolves.
Despite their promising results, most methods primarily focus on designing sample-importance criteria, i.e., deciding \textbf{what to select}.
The selected data volume, i.e., \textbf{how much to select} at each training stage, is usually fixed by the target ratio and treated merely as a budget constraint.
Thus, existing methods are often dynamic in sample identity but static in data volume, overlooking temporal data exposure as a control dimension for training dynamics.

In this work, we revisit data selection from an optimization perspective.
Our analysis shows that selected-data training induces an implicit regularization effect (Prop.~\ref{prop}), whose strength is modulated by the instantaneous selection ratio.
This offers a principled perspective on why reduced-data training can sometimes match or even outperform full-data training, and more importantly, identifies the selected data volume not simply as a static cost-control knob, but as a temporal control variable for regulating training dynamics. 
It further reveals a key trade-off: lower ratios strengthen selection-induced regularization and reduce cost but may bias and destabilize optimization, while higher ratios improve optimization fidelity but weaken the regularization benefit and reduce training efficiency.
However, a fixed ratio is inherently limited in its ability to exploit such regularization while preserving optimization stability.
This motivates scheduling the selection ratio across training as an additional data-volume dimension for improving generalization under the same overall budget, which presents a promising yet underexplored direction.
Thus a pivotal question is arised:
\textit{Beyond what to select, can data selection be improved by dynamically scheduling how much to select over training?}
% More importantly, this suggests that the selected data volume is not merely a static cost-control knob but also a temporal control variable for regulating training dynamics, revealing a key trade-off: lower ratios strengthen selection-induced regularization but may bias and destabilize optimization, while higher ratios improve optimization but weaken regularization.
% However, a fixed ratio cannot fully exploit this trade-off across different training stages.

\begin{wraptable}[22]{r}{0.7\linewidth}
    \centering
    \vspace{-.5cm}
\includegraphics[width=1.\linewidth]{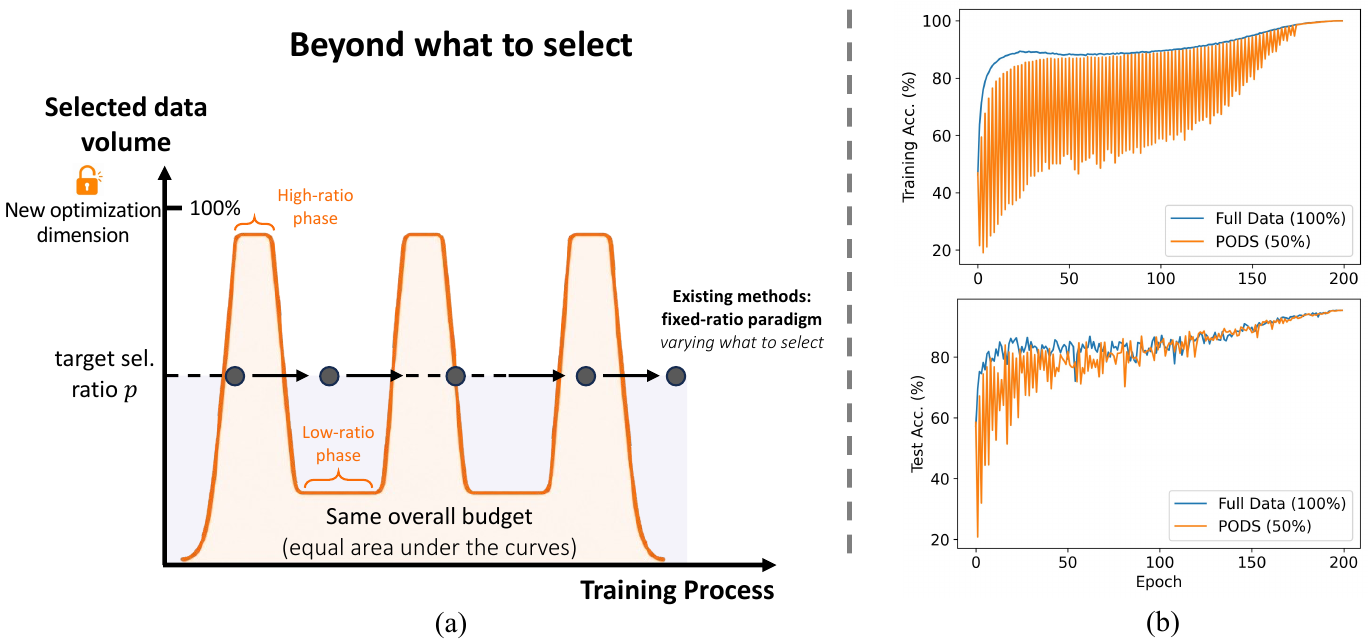}
    \vspace{-.4cm}
	\captionof{figure}{(a) \textbf{Oscillatory Data-volume Scheduling Mechanism.} Conventional data selection methods typically use a fixed ratio throughout training and mainly focus on what to select. In contrast, \model{} introduces an additional data-volume dimension by dynamically scheduling how much to select under the same cumulative budget. (b) \textbf{Visualization of Training Dynamics.} \model{} induces oscillatory training dynamics through scheduled data-volume variation, while the test accuracy steadily improves over training with only 50\% of the training budget, showing a favorable efficiency-generalization trade-off.
    % , the training accuracy exhibits oscillatory behavior induced by the scheduled data-volume variation, while the test accuracy steadily improves over training, suggesting improved generalization. 
    }
	\label{fig:reg-strength}
\end{wraptable}
% Fig.~\ref{fig-convergence-analysis} visualizes the optimization trajectories on CIFAR-10 using R-18.
% Under our method, the training accuracy shows clear oscillatory behavior, while the test accuracy remains more stable and steadily improves over training.
% Compared with the full dataset training, the proposed method periodically interrupts aggressive fitting of the training data, particularly in the low-ratio phases, yet does not destabilize the test trajectory. 
% As a result, the test accuracy matches or slightly exceeds the full-data baseline despite using only 50\% of the training data. 
% Thus, the results illustrate that the proposed oscillatory mechanism yields a more favorable optimization-generalization trade-off.
To answer this question, we propose \textbf{\model{}}, a Plug-and-play Oscillatory Data-volume Scheduling framework for efficient model training.
Rather than introducing another sample-scoring rule, \model{} dynamically modulates the instantaneous selected data volume.
Specifically, it alternates between high- and low-ratio phases: low-ratio phases strengthen selection-induced regularization, whereas high-ratio phases provide broader data coverage and stabilize optimization.
The duration of the low-ratio phases is automatically determined to increase regularization exposure while meeting the target selection-ratio budget.
In this way, the selected data volume is transformed from a static cost-control parameter into a temporal control signal for shaping training dynamics.
Importantly, since \model{} operates on an orthogonal dimension, it is naturally compatible with different sample-importance metrics, highlighting its flexibility as a general data-volume scheduling framework.
In this work, we use a simple and lightweight loss-based hard mining strategy to prioritize under-optimized samples by default. 
Moreover, its task-agnostic design makes PODS a lightweight, plug-and-play scheduler that can be applied across diverse training scenarios with negligible overhead.
Extensive experiments across diverse datasets, architectures, and tasks demonstrate that \model{} is an effective and broadly applicable framework for data-efficient learning.
On image classification, \model{} substantially reduces training cost while matching or surpassing full-data performance.
Its gains further extend to more challenging scenarios, including fine-grained, long-tailed, and out-of-distribution recognition, where \model{} maintains strong generalization with reduced training budgets.
Beyond classification, \model{} scales to more complex and resource-intensive regimes, including object detection and large-scale LLM instruction tuning, translating into practical savings of tens to hundreds of GPU hours.
For instance, Qwen-2.5-7B~\cite{qwen25}, \model{} matches full-data performance on MMLU~\cite{mmlu} and BBH~\cite{bbh} using only 50\% of the instruction-tuning budget.
Its effectiveness also generalizes across diverse architectures, including ResNet~\cite{resnet}, YOLOv8~\cite{yoco}, RT-DETR~\cite{rt-detr}, Qwen 2.5/3 series~\cite{qwen25} and Llama 3 series~\cite{llama}. 
Notably, these gains come with negligible additional overhead, e.g., adding only 5.4 seconds to a complete ImageNet-1k training process that requires nearly 100 GPU-hours.
% Meanwhile, these results also highlight its strong cross-architecture generalization, spanning across ResNet~\cite{resnet}, ViT~\cite{vit}, YOLOv8~\cite{yoco}, RT-DETR~\cite{rt-detr}, Qwen 2.5/3 series~\cite{qwen25} and Llama 3.1/3.2 series~\cite{llama}. 
% Importantly, these gains are achieved with high efficiency, e.g., adding only 5.4 seconds to a complete ImageNet-1k training process.
% consistently reduces training cost while preserving or improving generalization.
% On image classification, covering both coarse- and fine-grained datasets, \model{} achieves substantial training acceleration while matching or surpassing full-data performance.
% Its gains extend to more challenging scenarios, including long-tailed recognition and out-of-distribution generalization.
% Beyond classification, \model{} scales to object detection and large-scale LLM instruction tuning, achieving comparable or better performance with reduced training costs. 
% These results also demonstrate its cross-architecture generalization, including ResNet~\cite{resnet}, ViT~\cite{vit}, YOLOv8~\cite{yoco}, RT-DETR~\cite{rt-detr}, Qwen 2.5/3 series~\cite{qwen25} and Llama 3.1/3.2 series~\cite{llama}. 
% For example, with Qwen-2.5-7B, \model{} matches full-data performance on MMLU and BBH using only 50\% of the instruction-tuning budget.
To summarize, our main contributions are:
\textbf{(1)} We reveal temporal data-volume scheduling as an overlooked dimension in data selection. Beyond deciding what to select, we show that dynamically controlling how much to select modulates selection-induced implicit regularization and provides a new lever for improving generalization.
\textbf{(2)} We propose \textbf{\model{}}, a plug-and-play oscillatory data-volume scheduling framework for efficient model training. 
By scheduling the selected data volume over training, \model{} turns the fixed selection budget into a temporal control signal, making it lightweight, task-agnostic, and compatible with existing static and dynamic selection methods.
% (3) Due to its lightweight and task-agnostic design, \model{} scales across diverse tasks and can be combined with existing static and dynamic selection methods to further improve their performance with high efficiency.
\textbf{(3)} Experiments across 21 benchmarks spanning diverse tasks, architectures, and training regimes demonstrate that \model{} consistently improves the efficiency-generalization trade-off, e.g., improving ImageNet-1k accuracy by 0.4\% while reducing training cost by 50\%, and accelerating large-scale LLM instruction tuning by about $2\times$ without performance degradation.

% enhances training acceleration with high efficiency, e.g., improving accuracy by 0.4\% on ImageNet-1k while reducing 50\% training costs, and accelerating Qwen2.5-7B-Instruct and Llama3.2-3B instruction tuning by over $2\times$.

\section{Related Work}
% Data selection aims to reduce training costs by selecting informative subsets without sacrificing performance.
Existing methods can be broadly divided into static and dynamic data selection.
Static methods determine a subset before training, while dynamic methods update it during training.
Despite the differences, most existing works focus on designing metrics to estimate sample importance.

\subsection{Static Data Selection}
Static methods typically rely on predefined metrics to evaluate sample importance.
These criteria can be broadly categorized into geometry-based~\cite{moderate,ccs}, uncertainty-based~\cite{clip-selection,score-based-3}, and optimization-based approaches~\cite{dataset_pruning,glister}.
Geometry-based methods~\cite{herding,ccs,moderate,k-center-selection,sener2018activelearningconvolutionalneural} construct a coreset by improving coverage of the full dataset in feature space.
For instance, Herding~\cite{herding} selects samples with closer distances to the corresponding category centers.
CCS~\cite{ccs} measures the coverage of a dataset on a specific distribution by extending the geometric set cover problem to a distribution cover problem.
Moderate~\cite{moderate} selects samples that are closer to the median score.
Moreover, some methods also exploit local geometric structure.
D2~\cite{d2} evaluates sample importance through neighborhood relationships, while some methods~\cite{k-center-selection,active-learning1} combine geometric coverage objectives such as $k$-center selection with uncertainty-based criteria. 

Another line of work estimates sample importance from difficulty, uncertainty, or representativeness~\cite{dynamic_pruning-2,DUAL,data_diet,forgetting,rho-loss,clip-selection,moso,yang2025dynamic}.
EL2N and GraNd~\cite{data_diet} use the $\ell_2$-norm of the prediction error and the gradient norm as sample scores.
Forgetting~\cite{forgetting} quantifies how often a sample transitions from correct to incorrect prediction during training.
RHO-loss~\cite{rho-loss} selects samples that approximately maximize the reduction in generalization loss.
Other methods measure semantic representativeness using pretrained models, such as CLIP-based selection~\cite{clip-selection}, or characterize difficulty through geometric sparsity in feature space~\cite{yang2025dynamic}.
DUAL~\cite{DUAL} combines difficulty and prediction uncertainty to identify informative samples early in training.
% The work~\cite{clip-selection} evaluates samples' representativeness by assessing their semantic alignment between visual and textual features using CLIP.
MoSo~\cite{moso} determines sample importance by assessing its impact on the optimal empirical risk.
% The work~\cite{yang2025dynamic} defines sample difficulty as the geometric density in feature space and prioritizes sparse ones for training.
% DUAL~\cite{DUAL} identifies important samples from the early training stage based on both difficulty and prediction uncertainty.
Optimization-based approaches formulate data selection as a subset search or policy optimization problem. 
They estimate subset utility using influence functions~\cite{dataset_pruning}, bi-level optimization~\cite{glister}, temporal dual-depth scoring of sample difficulty~\cite{tdds}, and reinforcement learning~\cite{rl-selector}. 
To enhance diversity and coverage, some methods further incorporate facility location objectives~\cite{crest},  submodularity formulations~\cite{cgscore,submodular}, or gradient matching criteria~\cite{opt-based-3}.

\subsection{Dynamic Data Selection}
Unlike static selection, dynamic data selection updates the selected subset during training based on the model's evolving state~\cite{rs,rcap,zhou2026rethinking,dataagent}.
To avoid introducing huge computational costs into online training, existing dynamic methods typically leverage lightweight metrics for selection, while the selection ratio is kept fixed throughout training.
For example, the work~\cite{dynamic_pruning} introduces Upper Confidence Bound (UCB) and $\epsilon$-greedy metrics, leveraging uncertainty sampling and moving averages to dynamically score and select highly informative samples.
RCAP~\cite{rcap} proposes a class-aware, probabilistic sampling method that prioritizes difficult samples at a class-wise level, emphasizing more samples from hard categories. 
Similarly, Infobatch~\cite{infobatch} introduces a soft pruning mechanism to prune less informative samples according to their evolving loss distribution, achieving training acceleration with minimal performance degradation.
Beyond standard supervised learning, dynamic selection has been extended to other tasks. SAS~\cite{data-efficient-contrastive-ssl} studies sample contribution in contrastive self-supervised learning and develops a more data-efficient training framework.
Greats~\cite{greats} applies a greedy algorithm to improve batch quality based on a Taylor-expansion approximation.
Moreover, the work~\cite{yang2025dynamic} integrates data selection with data augmentation, prioritizing samples that are beneficial for augmentations.
\section{The Proposed Method}
\subsection{Problem Formulation: Dynamic Selection Beyond Fixed Ratios}
Let $\mathcal{D}$ denote the full training dataset of size $N$, and $\theta$ denote the model parameters. 
Conventional dynamic data selection constructs a sequence of subsets: $\mathcal{S}_1, \mathcal{S}_2,...,\mathcal{S}_T$, with a fixed cardinality $|S_t|=pN$, where $p\in (0,1]$ denotes the target selection ratio, $T$ is the total number of training epochs, and $t \in [1,T]$.
In this sense, these methods adapt which samples are selected during training, but usually do not adapt how much data is selected at different stages.

In contrast, we treat the selection ratio itself as a time-varying variable and construct subsets with cardinality $|S_t|=p_tN$, where $p_t \in (0,1)$ denotes the instantaneous selection ratio at time $t$.
Importantly, we impose a cumulative budget constraint based on the fixed-ratio baselines:
\begin{equation} \label{eq:budget-constraint}
    \frac{1}{T}\sum_{t=1}^T p_t N \leq pN \implies \frac{1}{T}\sum_{t=1}^T p_t \leq p.
\end{equation}
This constraint guarantees that the total sample usage, and hence the overall forward-pass budget, does not exceed that of the vanilla fixed-ratio selection setting.

\subsection{Selection Ratio Controls Implicit Regularization}
\begin{wrapfigure}[15]{r}{0.55\linewidth}
    \centering
    \vspace{-.3cm}
\includegraphics[width=1.\linewidth]{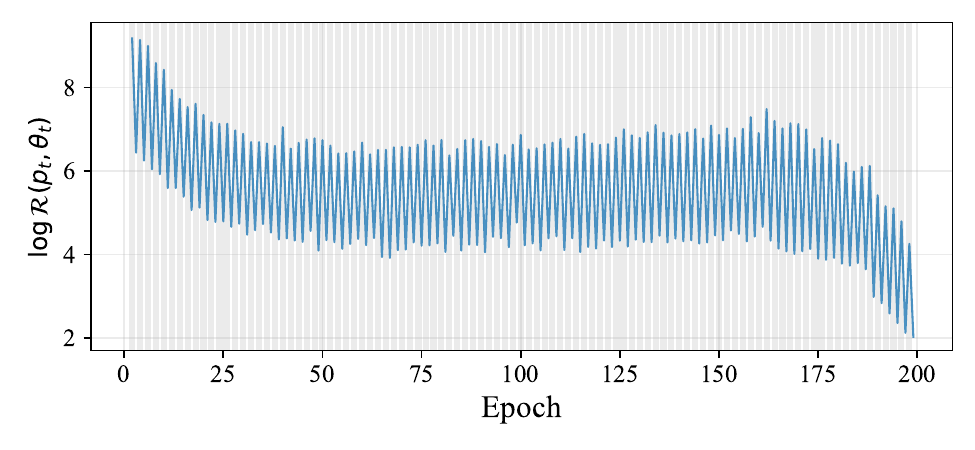}
    \vspace{-.5cm}
	\caption{Evolution of the regularization term $\mathcal{R}(p_t, \theta_t)$ under a 50\% target selection ratio. $\mathcal{R}$ follows a phase-aligned oscillatory pattern, increasing in low-ratio phases (gray shadowed) and decreasing in high-ratio phases.}
	\label{fig:reg-strength}
\end{wrapfigure}
We analyze how the selection ratio affects SGD dynamics. Our goal is to isolate the role of the time-varying ratio $p_t$ and show that subsampling induces a curvature-aware implicit regularization effect whose strength is directly controlled by the selected data volume.
This provides a principled explanation for why, under appropriate budgets, training on reduced data can generalize as well as, or even better than, full-dataset training. 
% To formalize the optimization dynamics induced by the oscillatory data selection mechanism, we analyze how varying the instantaneous data volume affects the expected SGD update. 

Let the empirical risk over the full dataset $\mathcal{D}$ of size $N$ be $L(\theta)=\frac{1}{N} \sum_{i=1}^N \ell_i(\theta)$, where $\theta$ denotes the model parameters. Following~\cite{jin2017escapesaddlepointsefficiently,li2019ssrgd}, we assume that $L(\theta)$ is twice differentiable with a Lipschitz continuous Hessian $H(\theta)=\nabla^2L(\theta)$.
We consider a single SGD update:
\begin{equation}
    \theta_{t+1}=\theta_t-\eta \hat{g}_t\left(\theta_t\right),
\end{equation}
where $\hat{g}_t\left(\theta_t\right)$ is the stochastic gradient computed from the selected subset at step $t$, and $\eta$ is the learning rate.
To characterize the effect of data selection on SGD dynamics, we study the expected loss after one update using a second-order Taylor expansion. This leads to the following proposition.
\begin{proposition}\label{prop}
    (Implicit Regularization via Data Selection). 
    The expected loss at the next iteration can be approximated as the deterministic gradient descent step plus an implicit regularization term $\mathcal{R}(p_t,\theta_t)$:
    \begin{equation}
    \mathbb{E}\left[L\left(\theta_{t+1}\right)\right] \approx L\left(\theta_t\right)-\eta\left\|\nabla L\left(\theta_t\right)\right\|^2+\frac{\eta^2}{2} \nabla L\left(\theta_t\right)^T H\left(\theta_t\right) \nabla L\left(\theta_t\right)+ \mathcal{R}(p_t,\theta_t),
    \end{equation}
    where $\mathcal{R}(p_t, \theta_t) = \frac{\eta^2}{2N} \left( \frac{1-p_t}{p_t} \right) \text{Tr}(H(\theta_t) C(\theta_t))$.
    Here, $C(\theta_t)$ denotes the empirical covariance of per-sample gradients, and the factor $\frac{1-p_t}{p_t}$ captures the volumetric modulation induced by selection.
\end{proposition}
The full proof is provided in Appendix~\ref{appendix-sec-proof}.

\begin{corollary}\label{corollary1}
    (Monotonicity of the Volumetric Regularization Strength). 
    Under uniform random sampling, suppose that the geometry term $\mathrm{Tr}(H(\theta_t)C(\theta_t))$ remains locally stable with respect to the sampling ratio. Then the magnitude of the implicit regularization term is modulated by $\lambda(p_t) = \frac{1-p_t}{p_t}$, which is decreasing for $p_t\in(0,1)$.
    Consequently, smaller selection ratios induce stronger implicit regularization, whereas larger selection ratios weaken this effect. 
\end{corollary}
This corollary highlights the sampling ratio as a direct modulator of the regularization strength: decreasing the ratio amplifies the selection-induced regularization. 
However, stronger regularization is not always better.
Similar to standard regularization, overly strong regularization can hinder optimization, and excessively small ratios may destabilize training and degrade final performance.
This reveals a key trade-off: the objective is not to maximize the regularization term alone, but to exploit its benefit while preserving optimization stability.
Motivated by this, we propose an oscillatory selection mechanism that alternates between low-ratio regularization phases, which strengthen regularization, and high-ratio phases, which restore stable optimization, all under a fixed target selection ratio.
In this way, oscillatory selection provides a simple mechanism for dynamically balancing regularization and optimization during training.

\subsection{Oscillatory Data Selection}
Building on the theoretical analysis above, our goal is to exploit the regularization benefit of reduced-data training without compromising optimization. 
We therefore propose an oscillatory ratio schedule that alternates between low-ratio and high-ratio phases under a target data budget.

\textbf{Selection Scheduling.} 
We first describe how \model{} schedules the selection ratio over training.
Given a target selection ratio $p$, we construct a periodic ratio trajectory consisting of $k$ low-ratio epochs followed by one high-ratio recovery epoch: $\underbrace{p_{low}, \dots, p_{low}}_{k \text{ epochs}}, p_{high}$.
Here, $k$ denotes the duration of the low-data phase, and each oscillation period has length $\tau=k+1$.
Within each period, the model is trained for $k$ consecutive epochs with ratio $p_{low}$, followed by one recovery epoch with ratio $p_{high}$.

To ensure that the cumulative number of forward passes does not exceed the prescribed target ratio $p$, the average ratio over each period should satisfy:
\begin{equation}\label{eq:budget_constraint}
    \frac{1}{k+1} \sum_{i=1}^{k+1} p_i \le p \implies \frac{k \cdot p_{low} + p_{high}}{k+1} \le p.
\end{equation}
Assuming $p_{low}<p<p_{high}$, this constraint gives $k \ge \frac{p_{high}-p}{p-p_{low}}$.
We therefore choose the smallest feasible integer
% Under this constraint, we choose the smallest feasible $k$ so as to achieve the highest oscillation frequency, yielding 
\begin{equation}
    k^* = \left\lceil \frac{p_{high} - p}{p - p_{low}} \right\rceil,
\end{equation}
which satisfies the cumulative budget while keeping the regularization phases sufficiently frequent.
This yields a bounded sample usage: $\frac{1}{\tau}\sum_tp_tN \leq pN$.

We further parameterize the schedule with a small stability margin $\epsilon$.
Specifically, we set the high ratio to $p_{high}=1-\epsilon$ and the low ratio to $p_{low}=\epsilon$ to maximize the oscillation amplitude and exploit the regularization benefits of low-ratio training.
% To achieve a large oscillation amplitude while preserving optimization stability, we introduce a stability margin $\epsilon$ and set the recovery phase to $p_{high}=1-\epsilon$.
% We then choose $p_{low}$ as aggressively as possible under the budget constraint. 
% When the target average ratio $p$ is relatively small, we set $p_{low}=\epsilon$ to maximize the oscillation amplitude.

Nevertheless, for larger target ratios, using an extremely small $p_{low}$ would mathematically constrain the feasible average ratio to 50\% when $k=1$.
In this case, we set $k=1$ and center the oscillation around the target ratio by imposing $\frac{p_{low}+p_{high}}{2}=p$, which gives $p_{low}^* = 2p - p_{high}$.
% In that case, we instead center the oscillation around the target ratio by imposing $\frac{p_{low}+p_{high}}{2}=p$, which gives $p_{low}^* = 2p - p_{high}$.

% 逻辑是，random可以只关注p，但是正比于pHC，那关注HC也可以提升。
% \textbf{Dynamic Instantiation via Boundary Collision.}
% To effectively leverage the regularization effect while maintaining the stability of the optimization process, we maintain a large selection amplitude $p_{high}-p_{low}$ and impose a strict lower bound for $p_{low}$, preventing the statistical validity of the mini-batch gradient and representation collapse.

% Given the constraint $\frac{1}{T}\sum_tp_t \leq p$, we can modulate the low-data phase length $k$ to satisfy: $\mathbb{E}[p_i] = \frac{p_{high} + k \cdot p_{low}}{k + 1} \leq p$.
% However, when $p$ is large, maintaining $p_{low}$ as an extremely low value, e.g., 0.05, mathematically restricts the maximum achievable ratio to 0.5. 
% Therefore, we relax the lower bound by setting the constraint to an equality ($\frac{p_{low} + p_{high}}{2} = p$), and derive the $p_{low}$ as $p_{low}^* = 2p - p_{high}$.

Overall, the schedule is given by: 
\begin{equation}\label{eq:parameter-definition}
    p_{low}^* = \begin{cases} 1-p_{high}, & \text{if } p < 0.5 \\ 2p - p_{high}, & \text{if } p \ge 0.5 \end{cases}, \quad \quad k^* = \begin{cases} \left\lceil \frac{p_{high} - p}{p - p_{low}} \right\rceil, & \text{if } p < 0.5, \\ 1, & \text{if } p \ge 0.5. \end{cases}
\end{equation}
In practice, we fix $\epsilon=0.05$, and all remaining scheduling parameters are automatically determined by the target ratio $p$ without additional tuning.

The oscillatory selection schedule in Eq.~\eqref{eq:parameter-definition} determines how much data is selected at each epoch, and is independent of what samples are selected.
Thus, \model{} can be instantiated with different sample-importance criteria.
This highlights \model{} as a plug-and-play data-volume scheduler rather than a specific sample selector.
In this work, we simply adopt loss-based hard mining as a lightweight default instantiation, selecting the top-$p_t$ fraction of samples $S_t = \mathop{\arg\text{top-}k} \ \ell_i(\theta_t)$.
Other selection metrics can be plugged into the same oscillatory schedule without changing the ratio-control mechanism, allowing PODS to complement existing static and dynamic data selection methods.

\section{Experiment}
\subsection{Experiment Settings}
\textbf{Datasets and Architectures.} We evaluate \model{} across a wide range of datasets, architectures, and learning paradigms to examine its effectiveness and scalability.
For image classification, we use coarse-grained benchmarks including CIFAR-10/100~\cite{cifar100}, Tiny-ImageNet~\cite{tiny}, and ImageNet-1k~\cite{imagenet}, and fine-grained benchmarks including FGVC-Aircraft~\cite{oxford-aircraft}, Oxford-IIIT Pets~\cite{oxford-pets}, Stanford Cars~\cite{stanford-cars}, and Oxford Flowers~\cite{oxford-flower}. 
We further evaluate more challenging recognition settings, including long-tailed recognition on ImageNet-LT and Places-LT~\cite{long-tailed}, and OOD generalization on ImageNet-A/O~\cite{imagenet-a}, ImageNet-R~\cite{imagenet-r}, and ImageNet-Hard~\cite{imagenet-hard}.
Beyond classification, we apply \model{} on MS-COCO~\cite{mscoco} for object detection with YOLOv8~\cite{yolov8} and RT-DETR~\cite{rt-detr}.
We also extend its scalability to large-scale LLM instruction tuning by fine-tuning Qwen 2.5/3~\cite{qwen25} LLaMA 3 series models~\cite{llama}, covering both general instruction benchmarks such as MMLU~\cite{mmlu}, BBH~\cite{bbh}, and reasoning-oriented benchmarks including AIME24~\cite{aime24}, Olympiad~\cite{olympiad}, GPQA-Diamond~\cite{gpqa}, and MMLU-Pro~\cite{mmlupro}.
Due to the limited space, detailed implementations are provided in Appendix.

% To prove that our method scales well on large-scale LLM instruction tuning, we train Qwen 2.5/3 and Llama 3 series models on xxx and conduct evaluation on benchmarks including MMLU~\cite{mmlu} and BBH~\cite{bbh}.
% To assess cross-architecture generalization, we leverage ResNet~\cite{resnet} and ViT~\cite{vit} series for classification tasks, YOLOv8~\cite{yolov8} for detection, and Qwen 2.5/3 series models for LLM instruction tuning. 
\textbf{Baselines.} We compare our method with 17 state-of-the-art baselines, including Herding~\cite{herding}, EL2N/GraNd~\cite{data_diet}, Glister~\cite{glister}, Forgetting~\cite{forgetting}, Moderate-DS~\cite{moderate}, CLIP-Sel~\cite{clip-selection}, Self-sup. prototypes~\cite{beyond}, MoSo~\cite{moso}, DP~\cite{dataset_pruning}, RL-Selector~\cite{rl-selector}, DUAL~\cite{DUAL}, Random*, Dyn. Sel. Aug.~\cite{yang2025dynamic}, UCB, $\epsilon$-Greedy~\cite{dynamic_pruning}, and InfoBatch~\cite{infobatch}.

\begin{table*}[]
    \centering
    \caption{Accuracy (\%) comparison with state-of-the-art data selection methods. All methods are trained with ResNet-18 on CIFAR-10/100 and ResNet-50 on Tiny-ImageNet.
Random* denotes per-epoch random selection under the same selection ratio.
Some results are taken from~\cite{infobatch,DUAL}.\label{tab:comparison_experiment}}
	\resizebox{0.8\textwidth}{!}{
    \begin{tabular}{l ccc|ccc|ccc}
    \toprule
\multirow{2}{*}{Method}
& \multicolumn{3}{c}{CIFAR-10}
& \multicolumn{3}{c}{CIFAR-100}
& \multicolumn{3}{c}{Tiny-ImageNet} \\
\cmidrule(lr){2-4}
\cmidrule(lr){5-7}
\cmidrule(lr){8-10}
& 30\% & 50\% & 70\%
& 30\% & 50\% & 70\%
& 30\% & 50\% & 70\% \\
\midrule
Full Dataset
& \multicolumn{3}{c}{95.6}
& \multicolumn{3}{c}{78.2}
& \multicolumn{3}{c}{45.0} \\ \midrule
    
    Random &90.2&92.3&93.9&69.7&72.1&73.8&29.8&37.2&42.2 \\ 
    Herding~\cite{herding} &80.1 &88.0&92.2 &69.6&71.8&73.1&29.4&31.6&39.8 \\
    EL2N~\cite{data_diet} &91.6&95.0&95.2 &69.5&72.1&77.2 &26.6&37.1&44.0 \\
    GraNd~\cite{data_diet}&91.2&94.6&95.3 &68.8&71.4&74.6 &29.7&36.3&43.2 \\
    Glister~\cite{glister}& 90.9&94.0&95.2 &70.4&73.2&76.6 &30.1&39.5&43.9 \\
    Forgetting~\cite{forgetting} &91.7&94.1&94.7 &69.9&73.1&75.3 &28.7&33.0&41.2 \\
    Moderate-DS~\cite{moderate}& 91.5&94.1&95.2&70.2&73.4&77.3 &30.6&38.2&42.8 \\
    CLIP-Sel~\cite{clip-selection}& 91.9&94.5&95.0 &70.8&73.7&77.0 &31.7&40.0 &46.0 \\
    Self-sup. proto.~\cite{beyond} &91.0&94.0&95.2&70.0&71.7&76.8 &27.7&37.9&43.4 \\
    MoSo~\cite{moso}& 91.1&94.2&95.3 &70.9&73.6&77.5 &31.2&38.5&43.4 \\
    DP~\cite{dataset_pruning}&90.8&93.8&94.9 &-&73.1&77.2 &-&-&- \\
    RL-Selector~\cite{rl-selector}&91.8&-&95.4 &71.1&-&77.6 &31.1&-&44.5 \\
    DUAL~\cite{DUAL}&91.8&95.4&95.3 &66.4&74.6&77.4 &-&-&-\\
    Random*& 93.0&94.5&94.8 &74.4&75.3&77.3 &41.5&42.8&43.1 \\
    UCB~\cite{dynamic_pruning}& 93.9&94.7&95.3 &-&75.3&77.3 &-&-&-    \\
    $\epsilon$-Greedy~\cite{dynamic_pruning}&94.1&94.9&95.2 &-&74.8&76.4&-&-&- \\
    Dyn. Sel. Aug.~\cite{yang2025dynamic} &94.6 & 95.1 & 95.5 &75.9&77.6& 78.6&41.5&45.1&48.5 \\
    % RCAP~\cite{rcap} & 94.7 & - & 95.2 & 77.6 &- & 78.6 &-&-&- \\
    InfoBatch~\cite{infobatch}& 94.7&95.1&95.6 &76.5&78.1&78.2 &42.2&43.2&43.8 \\ 
    % RS2 w/r& &94.0&- &95.4 &75.1&-&78.4 &&& \\
    % RS2 w/o& &93.1&-&95.3 &72.7&-&75.3 &&& \\
    \hline
    \model{} &\textbf{95.1}&\textbf{95.4}&\textbf{95.9}&\textbf{78.4}&\textbf{79.1}&\textbf{79.4} & \textbf{44.5} & \textbf{46.9} & \textbf{49.0} \\
    \bottomrule 
    \end{tabular}}
    % \vspace{-1mm}
\end{table*}

\subsection{Comparison with State-of-the-art Methods}
\textbf{Performance Comparisons.} 
We compare \model{} with representative static and dynamic data selection methods on CIFAR-10/100 and Tiny-ImageNet.
For \model{}, the reported ratio denotes the cumulative data budget.
As shown in Table~\ref{tab:comparison_experiment}, \model{} consistently achieves the best performance across datasets and selection ratios.
Notably, at the 30\% selection ratio, \model{} outperforms the leading baseline by 1.9\% on CIFAR-100 and 2.3\% on Tiny-ImageNet.
In particular, on CIFAR-100, \model{} is the only one that surpasses full-dataset training using only 30\% of the data budget.
% the lossless performance compared to full-dataset training with a 30\% selection ratio.
At higher selection ratios, \model{} further improves performance over full-dataset training, thereby accelerating training.
These results suggest that the oscillatory selection mechanism not only reduces training costs but also improves generalization.
% Meanwhile, when the selection ratio exceeds 50\%, our method outperforms the full dataset, achieving lossless training acceleration.
% Notably, these gains are achieved with extremely high efficiency due to the lightweight design of the proposed method.

\textbf{Efficiency Comparisons.}
Following~\cite{infobatch}, we evaluate both accuracy and efficiency on large-scale ImageNet-1k.
As shown in Table~\ref{tab:imagenet-1k}, \model{} achieves the best overall accuracy-efficiency trade-off, delivering the highest accuracy while incurring nearly zero additional overhead.
Compared with full-data training, \model{} improves accuracy by 0.4\% and reduces the total cost from 140 to 84 GPU hours, saving 56 GPU hours.
This advantage is particularly important at scale. 
Although many static selection methods reduce the number of training samples, they often require expensive preprocessing, pretraining, or scoring stages, which, in several cases, are even higher than those of full-data training.
In contrast, \model{} avoids additional pretraining or complex metric computation: it only \textbf{adds only 5.4 seconds of overhead} over the entire 90-epoch ImageNet-1k training process. 
These results demonstrate that \model{} offers a practical accuracy-efficiency trade-off for large-scale training.
% without relying on expensive preprocessing, auxiliary optimization, or complex metric computation.
\begin{table*}[]
       \centering
    \caption{Results on ImageNet-1k at a 60\% selection ratio using ResNet-50 on an 8-A100 server. We report wall-clock training time (h) and total GPU hours. Glister and CG-Score are not reported due to high computational and memory costs~\cite{moderate}. Some results are taken from~\cite{infobatch,yang2025dynamic}.}\label{tab:imagenet-1k}
	\resizebox{0.97\textwidth}{!}{
    \begin{tabular}{c|cccccccccccccc}
    \toprule[1.pt]
    Method &Herding&EL2N&GraNd&Forgetting&RL-Selector&SSP&Moderate&CLIP-Sel&DUAL&UCB&Infobatch &Ours&Whole Dataset \\ \hline
    Acc. (\%) &71.1&72.3&71.0&72.5& 73.4&70.0&73.1&73.2&75.5&75.8&76.5 & \textbf{76.8} &76.4 \\ \hline
    Time (h) &10.5&10.5&10.5&10.5&10.5&10.5&10.5&10.5&10.5&10.5& 10.5&10.5&17.5 \\
    Overhead (h) &$>$17.5&$>$17.5&$>$17.5&$>$17.5&$>$17.5&$>$24.0&$>$17.5&$>$1.6&>11.6&0.03&0.0028& \textbf{0.0015} &0.0 \\
    Total GPU Hours& $>$224.0 &$>$224.0&$>$224.0&$>$224.0&$>$224.0&$>$276.0&$>$224.0&$>$96.8&>176.8&\textbf{84.0}& \textbf{84.0}& \textbf{84.0} &140.0 \\
    \bottomrule[1.pt]
    \end{tabular}}
\end{table*}

\begin{figure*}[]
    \centering
    \includegraphics[width=.99\linewidth]{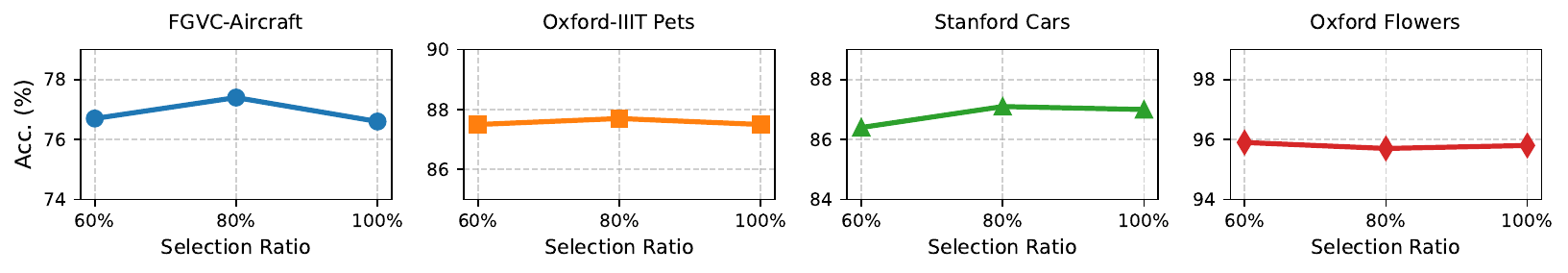}
	\caption{Results on four fine-grained recognition benchmarks using ResNet-50.}
	\label{fig:fine-grained-datasets}
    \vspace{-5pt}
\end{figure*}

%  \begin{table*}[t]
%  \caption{Top-1 classification accuracy (\%) on ImageNet-LT and Places-LT~\cite{long-tailed} at a 50\% selection ratio. We report wall-clock GPU hours on a 4-V100 server. }
% 		\label{tab:long-tailed}
%     \centering
%     \setlength{\tabcolsep}{2.5pt}
%         \resizebox{.7\textwidth}{!}{\begin{tabular}{l|cccc|cccc|c}
%           \bottomrule[1.1pt]
%          &\multicolumn{4}{c}{ImageNet-LT} &\multicolumn{4}{|c|}{Places-LT} & \multirow{2}{*}{Costs (h)}\\ \cline{2-9}
%         & \textbf{Many} & \textbf{Med.} & \textbf{Few} & \textbf{Acc.} &  \textbf{Many} & \textbf{Med.} & \textbf{Few} & \textbf{Acc.}&  \\ \hline
%       Full Dataset  &43.2 & 35.1 & 18.5 & 35.6 & \textbf{44.7}  & 37.0 & 25.3 & 35.9 & 35.2 \\ 
%          \model{} & \textbf{43.0} &\textbf{35.5}&\textbf{20.8}&\textbf{36.3}&  43.0  &  \textbf{40.8}  & \textbf{29.4} &  \textbf{39.2} & \textbf{16.2} \\ 
%         % Full Dataset  & \textbf{44.7}  & 37.0 & 25.3 & 35.9  &44.6 &36.8& 25.2& 46.4 & \\
%         % Ours & 43.8  &  40.8  & 28.9 &  39.4  & 43.7 & 40.5&28.4& 50.2 \\ 
%          \bottomrule[1.1pt]
%         \end{tabular}}
% \end{table*}

\begin{table*}[t]
\centering
% \caption{
% Generalization results on long-tailed classification and object detection.
% Left: Top-1 accuracy (\%) on ImageNet-LT and Places-LT~\cite{long-tailed} at a 50\% selection ratio, with wall-clock GPU hours measured on a 4-V100 server.
% Right: MS-COCO~\cite{mscoco} object detection results using YOLOv8~\cite{yolov8} and RT-DETR~\cite{rt-detr} under different selection ratios. We report mAP (\%).
% }
% \label{tab:lt_detection}
\setlength{\tabcolsep}{3pt}
\renewcommand{\arraystretch}{1.05}

\begin{minipage}[t]{0.62\textwidth}
\centering
\captionof{table}{Long-tailed classification accuracy (\%) on ImageNet-LT and Places-LT~\cite{long-tailed} at a 50\% selection ratio. We report wall-clock GPU hours on a 4-V100 server.}
\vspace{-2pt}
\label{tab:long-tailed}
\resizebox{\linewidth}{!}{
\begin{tabular}{l|cccc|cccc|c}
\toprule
\multirow{2}{*}{Method}
& \multicolumn{4}{c|}{ImageNet-LT}
& \multicolumn{4}{c|}{Places-LT}
& \multirow{2}{*}{Cost} \\
\cline{2-9}
& Many & Med. & Few & Acc.
& Many & Med. & Few & Acc.
&  \\
\midrule
Full Dataset
& \textbf{43.2} & 35.1 & 18.5 & 35.6
& \textbf{44.7} & 37.0 & 25.3 & 35.9
& 35.2h \\
\model{}
& 43.0 & \textbf{35.5} & \textbf{20.8} & \textbf{36.3}
& 43.0 & \textbf{40.8} & \textbf{29.4} & \textbf{39.2}
& \textbf{16.2h} \\
\bottomrule
\end{tabular}
}
\end{minipage}
\hfill
% \vspace{3pt}
\begin{minipage}[t]{0.35\textwidth}
\centering
\captionof{table}{Object detection mAP (\%) on MS-COCO~\cite{mscoco} using YOLOv8~\cite{yolov8} and RT-DETR~\cite{rt-detr}.}
\label{tab:detection}
\resizebox{\linewidth}{!}{
\begin{tabular}{l|cccc}
\toprule
Method & 70\% & 80\% & 90\% & Full \\
\midrule
YOLOv8 & 38.5 & 39.1 & \textbf{39.8} & 39.6 \\
RT-DETR & 55.9 & 56.6 & \textbf{57.6} & 57.5 \\
\bottomrule
\end{tabular}
}
\end{minipage}
\vspace{-5pt}
\end{table*}
\subsection{Generalization Analysis}
\textbf{Generalization to Fine-Grained Classification.}
While many methods do not evaluate on fine-grained recognition, we evaluate \model{} on four fine-grained benchmarks, including FGVC-Aircraft~\cite{oxford-aircraft}, Oxford-IIT Pets~\cite{oxford-pets}, Stanford Cars~\cite{stanford-cars}, and Oxford Flowers~\cite{oxford-flower}.
As shown in Fig.~\ref{fig:fine-grained-datasets}, \model{} consistently preserves or improves full-data performance with reduced training costs.
For instance, at an 80\% selection ratio, \model{} achieves the best results and often surpasses full-dataset training.
This is encouraging for fine-grained recognition, where models rely on subtle discriminative cues and are sensitive to redundant or less informative samples.
These results suggest that the oscillatory scheduling can improve training efficiency while maintaining fine-grained generalization.

\textbf{Generalization to Long-tailed Classification.}
 We further evaluate \model{} on ImageNet-LT and Places-LT to assess its robustness under challenging imbalanced data distributions.
As shown in Table~\ref{tab:long-tailed}, \model{} surpasses full-data training on both datasets using only 50\% of the data budget.
The improvement mainly comes from medium- and few-shot categories: \model{} improves few-shot accuracy from 18.5\% to 20.8\% on ImageNet-LT, and improves medium- and few-shot accuracy by 3.8\% and 4.1\% on Places-LT, respectively.
Notably, these gains are achieved with reduced training costs, highlighting the effectiveness of \model{}.
% These results show that \model{} effectively helps mitigate head-class dominance while preserving sufficient sample coverage.

\textbf{Generalization to Object Detection.}
To evaluate the cross-task generalization of \model{}, we further apply it to MS-COCO object detection using YOLOv8 and RT-DETR.
As shown in Table~\ref{tab:detection}, \model{} maintains competitive detection performance under reduced data budgets.
These results suggest that \model{} generalizes well to more complex detection training with more complex objectives and different detector architectures.

% \begin{table}[]
%     \centering
%     \caption{Results on MS-COCO~\cite{mscoco} using YOLOv8~\cite{yolov8} and RT-DETR~\cite{rt-detr} across different selection ratios. We report mAP (\%). \label{tab:detection}}
% 	\resizebox{0.5\textwidth}{!}{
%     \begin{tabular}{c|cccc }
%     \bottomrule[1.2pt]
%     &70\% & 80\% & 90\% & Full Dataset \\ \hline
%      % Random*  &37.5 & 37.9 &38.5& - \\
%      YOLOv8 & 38.5 & 39.1 & 39.8& 39.6 \\
%      RT-DETR & 55.9 & 56.6 & 57.6 & 57.5 \\
%     \bottomrule[1.2pt]
%     \end{tabular}}
% \end{table}

\begin{table}[]
\centering
\caption{
LLM instruction-tuning results on BBH~\cite{bbh} and MMLU~\cite{mmlu} using Qwen-2.5-7B-Instruct and LLaMA-3.2-3B.
Training cost is reported as wall-clock time on an 8-A100 server.
}
\label{tab:llm}
\setlength{\tabcolsep}{4.8pt}
\renewcommand{\arraystretch}{1.08}
\resizebox{0.99\textwidth}{!}{
\begin{tabular}{l l c | c c c c | c c}
\toprule
Model & Method & Selection Ratio & BBH & MMLU & Avg. & $\Delta$ Avg. & Cost (h) & Speedup \\
\midrule
\multirow{2}{*}{Qwen-2.5-7B-Instruct}
& Full Data & 100\% & 66.2 & 74.0 & 70.1 & -- & 32.8 & -- \\
& \model{} & 50\% & 66.2 & 74.0 & 70.1 & 0.0 & 16.9 & $1.9\times$ \\
\midrule
\multirow{2}{*}{LLaMA-3.2-3B}
& Full Data & 100\% & 48.4 & 55.9 & 52.2 & -- & 17.6 & -- \\
& \model{} & 50\% & 48.4 & 56.0 & 52.2 & 0.0 & 9.0 & $2.0\times$ \\
\bottomrule
\end{tabular}}
\centering
\caption{Reasoning-oriented LLM fine-tuning results using DAPO-MATH-17K~\cite{dapo} on mathematical reasoning and general capability benchmarks with the same optimization steps.
}
\label{tab:llm_reasoning}
\setlength{\tabcolsep}{4.2pt}
\renewcommand{\arraystretch}{1.08}
\resizebox{0.98\linewidth}{!}{
\begin{tabular}{l l c | cc | cc | c}
\toprule
\multirow{2}{*}{Model} 
& \multirow{2}{*}{Method}
& \multirow{2}{*}{Selection Ratio}
& \multicolumn{2}{c|}{Mathematical Reasoning} 
& \multicolumn{2}{c|}{General Capability} 
& \multirow{2}{*}{Avg.} \\
\cline{4-7}
& & 
& AIME24 
& Olympiad 
& GPQA-Diamond 
& MMLU-Pro 
&  \\
\midrule
\multirow{2}{*}{Qwen3-4B}
& Full Data & 100\% 
& 50.0 & 65.7 & 35.8 & 61.1 & 53.2 \\
& \model{} & 50\% 
& 50.0 & \textbf{66.0} & \textbf{35.9} & \textbf{61.5} & \textbf{53.4} \\
\midrule
\multirow{2}{*}{Qwen3-8B}
& Full Data & 100\% 
& 53.3 & 64.5 & 36.4 & 65.8 & 55.0 \\
& \model{} & 50\% 
& 53.3 & \textbf{65.1} & \textbf{37.4} & \textbf{66.4} & \textbf{55.6} \\
\bottomrule
\end{tabular}
}
\vspace{-5pt}
\end{table}

% \begin{table}[]
% \centering
% \caption{
% Step-matched mathematical reasoning fine-tuning results on Qwen3-4B and Qwen3-8B.
% We report Pass@1 (\%) on mathematical reasoning and general capability benchmarks.
% }
% \label{tab:drop_ratio_results}
% \setlength{\tabcolsep}{5pt}
% \renewcommand{\arraystretch}{1.08}
% \resizebox{0.95\linewidth}{!}{
% \begin{tabular}{l c | cc | cc | c}
% \toprule
% \multirow{2}{*}{Model} 
% & \multirow{2}{*}{Data} 
% & \multicolumn{2}{c|}{Mathematical Reasoning} 
% & \multicolumn{2}{c|}{General Capability} 
% & \multirow{2}{*}{Avg.} \\
% \cline{3-6}
% & 
% & AIME24 
% & Olympiad 
% & GPQA-Diamond 
% & MMLU-Pro 
% &  \\
% \midrule
% \multirow{2}{*}{Qwen3-4B}
% & 50\% 
% & 50.0 & \textbf{66.0} & \textbf{35.9} & \textbf{61.5} & \textbf{53.4} \\
% & Full 
% & 50.0 & 65.7 & 35.8 & 61.1 & 53.2 \\
% \midrule
% \multirow{2}{*}{Qwen3-8B}
% & 50\% 
% & 53.3 & \textbf{65.1} & \textbf{37.4} & \textbf{66.4} & \textbf{55.6} \\
% & Full 
% & 53.3 & 64.5 & 36.4 & 65.8 & 55.0 \\
% \bottomrule
% \end{tabular}
% }
% \end{table}

\textbf{Generalization to LLM Fine-tuning.}
We evaluate \model{} on large-scale LLM instruction tuning to examine its scalability beyond vision tasks.
Following~\cite{greats}, we fine-tune Qwen-2.5-7B-Instruct and LLaMA-3.2-3B using instruction mixtures including FLAN V2~\cite{flanv2}, CoT~\cite{cot-training-set}, DOLLY~\cite{dolly}, and OPEN ASSISTANT~\cite{openassistant}, and evaluate them on MMLU and BBH.
As shown in Table~\ref{tab:llm}, \model{} matches full-data performance while using only half of the instruction-tuning budget, reducing wall-clock training time by about $2\times$.
This demonstrates that \model{} scales to LLM instruction tuning and provides practical training acceleration without degrading downstream capability.

We further study \model{} in more challenging reasoning-oriented fine-tuning settings.
Here, we keep the optimization horizon the same as full-data training and fine-tune Qwen3-4B/8B on DAPO-MATH-17K~\cite{dapo} for 1500 steps with a rollout batch size of 32, and evaluate them on AIME24, Olympiad Bench, GPQA-Diamond, and MMLU-Pro.
As shown in Table~\ref{tab:llm_reasoning}, \model{} consistently achieves higher average performance than full-data training under the same training budget. 
These results suggest that \model{} not only accelerates general instruction tuning but also improves reasoning generalization when the optimization horizon is fixed, highlighting the benefit of oscillatory data-volume scheduling. 

\textbf{Compatibility with Existing Selection Methods.}
\model{} is designed as a plug-and-play framework that can be applied on top of existing data selection methods.
To evaluate this, we integrate it with representative static and dynamic selection methods, including Glister, Moderate, UCB, $\epsilon$-Greedy, and InfoBatch.
As shown in Table~\ref{tab:scalability-other-metric}, adding \model{} consistently improves performance across selection ratios.
For example, at the 30\% selection ratio, \model{} improves Moderate and Glister by 7.0\% and 6.4\%, respectively. 
It also improves dynamic methods, with gains of 4.6\% for UCB and 3.7\% for $\epsilon$-Greedy at the 50\% ratio.
Meanwhile, these improvements are achieved with negligible additional overhead due to the lightweight design.
These results support that \model{} operates on a complementary dimension of data selection.
It modulates the selected data volume over training, enabling existing methods to better balance optimization and regularization under the same overall budget.

\begin{table}[]
    \begin{minipage}{0.48\textwidth}
        \centering
\caption{
Compatibility with existing static and dynamic data selection methods on CIFAR-100 using ResNet-18.
}
\label{tab:scalability-other-metric}
\resizebox{1\textwidth}{!}{
\begin{tabular}{lccc}
\toprule
Method & 30\% & 50\% & 70\% \\
\midrule
Moderate~\cite{moderate}
& 70.2 & 73.4 & 77.3 \\
\model{}+Moderate
& \textbf{77.2} {\small (+7.0)}
& \textbf{78.4} {\small (+5.0)}
& \textbf{79.2} {\small (+1.9)} \\
\cmidrule(lr){1-4}
Glister~\cite{glister}
& 70.4 & 73.2 & 76.6 \\
\model{}+Glister
& \textbf{76.8} {\small (+6.4)}
& \textbf{78.7} {\small (+5.5)}
& \textbf{79.1} {\small (+2.5)} \\
\midrule
UCB~\cite{dynamic_pruning}
& 74.0 & 75.3 & 77.3 \\
\model{}+UCB
& \textbf{77.7} {\small (+3.7)}
& \textbf{78.9} {\small (+4.6)}
& \textbf{79.3} {\small (+2.0)} \\
\cmidrule(lr){1-4}
$\epsilon$-Greedy~\cite{dynamic_pruning}
& 73.9 & 74.8 & 76.4 \\
\model{}+$\epsilon$-Greedy
& \textbf{77.3} {\small (+3.9)}
& \textbf{78.5} {\small (+3.7)}
& \textbf{78.7} {\small (+2.3)} \\
\cmidrule(lr){1-4}
InfoBatch~\cite{infobatch}
& 76.5 & 78.1 & 78.2 \\
\model{}+InfoBatch
& \textbf{78.5} {\small (+2.0)}
& \textbf{78.8} {\small (+0.6)}
& \textbf{79.3} {\small (+1.1)} \\
\bottomrule
\end{tabular}}
    \end{minipage}
    \hfill
\begin{minipage}{0.5\textwidth}
    \begin{minipage}{.95\textwidth}
        \centering
    \caption{Cross-optimizers accuracy on CIFAR-10 using R-18. \label{tab:optimizer}}
	\resizebox{.8\textwidth}{!}{
    \begin{tabular}{c|cccc }
    \toprule
     &30\% &40\% &50\% &Full dataset\\ \hline
    Adam &91.5 &91.9&\textbf{92.1}& 92.0 \\
    Lars &95.4&95.5&\textbf{95.6}&  95.5 \\
    Lamb &95.0&95.0&\textbf{95.1}& 95.0 \\
    SGD &95.0&95.4&\textbf{95.6}& \textbf{95.6}\\
    \bottomrule
    \end{tabular}}
    \end{minipage}

        \begin{minipage}{1\textwidth}
            \begin{minipage}{.46\textwidth}
                    \centering
                    \caption{Effect of different $\epsilon$ values on C-100 using R-18. \label{tab:ablation-p-value}}
                	\resizebox{1\textwidth}{!}{
                    \begin{tabular}{c|ccc}
                    \toprule
                    $\epsilon$ &  30\% &  50\% &70\% \\ \hline
                    0.01 & 78.5 & 79.3 & 79.4 \\
                    0.05 &78.4 & 79.1& 79.4\\
                    0.1 & 78.2 & 79.0 & 79.3 \\
                    0.2 & 78.1 & 78.7 & 79.0 \\
                    \bottomrule 
                    \end{tabular}}
            \end{minipage}
            \hfill
            \begin{minipage}{.49\textwidth}
                    \centering
                    \caption{Effect of oscillatory scheduling (OS) and hard mining (HM).
                    }
                    \label{tab:ablation-study}
                    \resizebox{1\textwidth}{!}{
                    \begin{tabular}{ccccc}
                    \toprule
                    OS & HM & 30\% & 50\% & 70\% \\
                    \midrule
                         &      & 41.5 & 42.8 & 43.1 \\
                    \checkmark &      & 42.0 & 45.9 & 47.8 \\
                         & \checkmark & 41.7 & 45.5 & 46.1 \\
                    \checkmark & \checkmark & \textbf{44.5} & \textbf{46.9} & \textbf{49.0} \\
                    \bottomrule
                    \end{tabular}}
            \end{minipage}
     
    \end{minipage}
 %    \begin{minipage}{1\textwidth}
 %         \centering
 %    \caption{Effect of different $\epsilon$ values on CIFAR-100 using ResNet-18. \label{tab:ablation-p-value}}
	% \resizebox{.6\textwidth}{!}{
 %    \begin{tabular}{c|ccc}
 %    \toprule
 %    $\epsilon$ &  30\% &  50\% &70\% \\ \hline
 %    0.01 & 78.5 & 79.3 & 79.4 \\
 %    0.05 &78.4 & 79.1& 79.4\\
 %    0.1 & 78.2 & 79.0 & 79.3 \\
 %    0.2 & 78.1 & 78.7 & 79.0 \\
 %    \bottomrule 
 %    \end{tabular}}
 %    \end{minipage}
\end{minipage}
    
\end{table}

\textbf{Cross-Optimizer Generalization.}
We further evaluate \model{} with different optimizers, including Adam, Lars, Lamb, and SGD. 
As shown in Table~\ref{tab:optimizer}, \model{} remains effective across different optimization algorithms and selection ratios.
At the 50\% selection ratio, it matches or slightly exceeds full-data training across optimizers.
These results suggest that oscillatory selection is not specific to SGD, but is broadly compatible with different optimizers.

% \begin{table}[]
%     \centering
%     \caption{Results across different optimizers on CIFAR-10 using R-18. We report top-1 accuracy (\%).\label{tab:optimizer}}
% 	\resizebox{0.45\textwidth}{!}{
%     \begin{tabular}{c|cccc }
%     \toprule
%      &30\% &40\% &50\% &Full dataset\\ \hline
%     Adam &91.5 &91.9&\textbf{92.1}& 92.0 \\
%     Lars &95.4&95.5&\textbf{95.6}&  95.5 \\
%     Lamb &95.0&95.0&\textbf{95.1}& 95.0 \\
%     SGD &95.0&95.4&\textbf{95.6}& \textbf{95.6}\\
%     \bottomrule
%     \end{tabular}}
% \end{table}

\textbf{Generalization to OOD Benchmarks.}
\begin{figure*}[t]
    \centering
    \includegraphics[width=.99\linewidth]{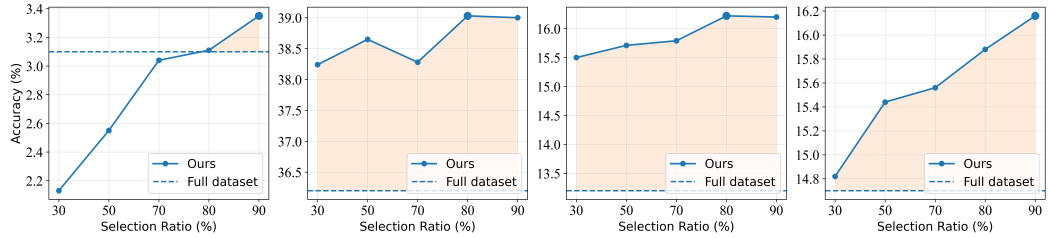}
	\caption{Out-of-distribution generalization on ImageNet-A/R/O/Hard. We report AUPR (\%) on ImageNet-O and accuracy (\%) on others.}
	\label{fig:ood-datasets}
\vspace{-5pt}
\end{figure*}
To evaluate the generalization of models trained with \model{} under distribution shifts, we evaluate them on ImageNet-A/R/O/Hard.
As shown in Fig.~\ref{fig:ood-datasets}, \model{} consistently achieves stronger OOD performance than full-data training across ratios.
These gains suggest that reduced-data training, when properly scheduled, can improve not only in-distribution accuracy but also generalization under distribution shifts.
This observation is consistent with our regularization perspective: the low-ratio phases may help reduce overfitting to the training distribution, while the high-ratio phases preserve sufficient data exposure and stabilize optimization.

\subsection{Ablation Study}
\textbf{Evaluation of Oscillation Margin $\epsilon$.}
\model{} uses a single oscillation margin $\epsilon$ to parameterize the scheduling range, from which the remaining parameters are automatically derived given the target selection ratio.
To assess its sensitivity, we vary $\epsilon$ while keeping the target selection ratio fixed.
As shown in Table~\ref{tab:ablation-p-value}, \model{} maintains stable performance across different $\epsilon$ values, highlighting its stability.
We also observe that larger $\epsilon$ values slightly reduce accuracy.
This is consistent with our analysis: increasing $\epsilon$ narrows the gap between low- and high-ratio phases, suppresses the oscillation amplitude, and weakens the selection-induced regularization effect.
% which is consistent with our analysis: a larger margin suppresses the oscillation amplitude, reduces the contrast between low- and high-phases, and weakens the induced regularization effect.

% \begin{table}[]
%     \centering
%     \caption{Effect of different components in our method on CIFAR-100 using R-18. \label{tab:ablation-study}}
% 	\resizebox{0.5\textwidth}{!}{
%     \begin{tabular}{cc|ccc}
%     \bottomrule[1.2pt]
%     Osc. Scheduling & Hard Mining & 30\% & 50\% & 70\% \\ \hline
%    %  & & 74.4 & 75.3 & 77.3 \\
%    % \checkmark & &77.9 & 78.9 & 79.2 \\
%    % & \checkmark & 78.1 & 78.7 & 79.0 \\
%    % \checkmark & \checkmark & \textbf{78.4} & \textbf{79.1} &  \textbf{79.4} \\
%      & & 41.5 & 42.8 & 43.1 \\
%    \checkmark & &42.0 & 45.9 & 47.8  \\
%    & \checkmark & 41.7 & 45.5 & 46.1 \\
%    \checkmark & \checkmark & \textbf{44.5} & \textbf{46.9} &  \textbf{49.0} \\
%     \bottomrule[1.2pt]
%     \end{tabular}}
% \end{table}
% \begin{wraptable}[8]{r}{0.32\linewidth}
% \vspace{-10pt}
% \centering
% \caption{
% Effect of oscillatory scheduling (OS) and hard mining (HM) on Tiny-IN1k.
% }
% \label{tab:ablation-study}
% % \setlength{\tabcolsep}{4.5pt}
% % \renewcommand{\arraystretch}{1.05}
% \resizebox{.3\textwidth}{!}{
% \begin{tabular}{ccccc}
% \toprule
% OS & HM & 30\% & 50\% & 70\% \\
% \midrule
%      &      & 41.5 & 42.8 & 43.1 \\
% \checkmark &      & 42.0 & 45.9 & 47.8 \\
%      & \checkmark & 41.7 & 45.5 & 46.1 \\
% \checkmark & \checkmark & \textbf{44.5} & \textbf{46.9} & \textbf{49.0} \\
% \bottomrule
% \end{tabular}}
% \end{wraptable}
\textbf{Evaluation of Different Components.}
As shown in Table~\ref{tab:ablation-study}, we evaluate key components in \model{} on Tiny-ImageNet using R-50 across selection ratios.
Both components improve performance, and their combination consistently achieves the best results.
This indicates that the data-volume dimension and the sample-scoring dimension are complementary: oscillatory scheduling improves how much data is selected over training, while hard mining provides a simple instantiation of which samples to select.

% When using either oscillatory selection or hard sample mining, the performance can be improved.
% This is because introducing the oscillatory selection mechanism introduces an implicit regularization penalty, whereas hard sample mining forces the network to focus on under-learned patterns, both contributing positively to the performance.
% More importantly, combining both components consistently yields the best results across all ratios.
% As a result, removing either component leads to a consistent drop in performance.

\section{Conclusion}
We present \model{}, a plug-and-play oscillatory data-volume scheduling framework that shifts data selection beyond the conventional focus on what samples to select. 
By identifying the selected data volume as a new optimization dimension, \model{} dynamically schedules how much data is exposed to the model over training, exploiting the regularization benefit while preserving optimization stability.
Thanks to its lightweight, task-agnostic design, \model{} scales across diverse tasks and architectures and can be readily integrated with existing static and dynamic selection methods with negligible additional overhead.
We hope this work provides the community with a broader view of data selection. Beyond estimating sample importance, data volume itself can be scheduled as a principled mechanism for developing data-efficient learning.

 %%%%%%%%%%%%%%%%%%%%%%%
\bibliography{ref}
\bibliographystyle{plain}

%%%%%%%%%%%%%%%%%%%%%%%%%%%%%%%%%%%%%%%%%%%%%%%%%%%%%%%%%%%%

% \appendix
% \input{appendix}

%%%%%%%%%%%%%%%%%%%%%%%%%%%%%%%%%%%%%%%%%%%%%%%%%%%%%%%%%%%%

% \newpage
% \input{checklist.tex}

\end{document}